\documentclass[sigconf]{acmart}

\usepackage{booktabs}
\usepackage{multirow}

\AtBeginDocument{%
  }

\copyrightyear{2026}
\acmYear{2026}
\setcopyright{cc}
\setcctype{by}
\acmConference[MM '26] {Proceedings of the 34th ACM International Conference on Multimedia}{November 10--14, 2026}{Rio de Janeiro, Brazil.}
\acmBooktitle{Proceedings of the 34th ACM International Conference on Multimedia (MM '26), November 10--14, 2026, Rio de Janeiro, Brazil}
\acmISBN{979-8-4007-2213-4/2026/11}
\acmDOI{10.1145/3767308.3837688}

\begin{document}

\title{Identity-Preserving Text-to-Video Generation via Agentic Enhancement and Semantic Repair}

\author{Jiayi Gao}
\email{gaojiayi26@stu.pku.edu.cn}
\affiliation{%
  \institution{Wangxuan Institute of Computer Technology, Peking University}
  \city{Beijing}
  \country{China}}

\author{Changcheng Hua}
\email{hcc@stu.pku.edu.cn}
\affiliation{%
  \institution{Wangxuan Institute of Computer Technology, Peking University}
  \city{Beijing}
  \country{China}}

\author{Jiaqi Tang}
\email{tangjiaqi@stu.pku.edu.cn}
\affiliation{%
  \institution{Wangxuan Institute of Computer Technology, Peking University}
  \city{Beijing}
  \country{China}}

\author{Yuxin Peng}
\email{pengyuxin@pku.edu.cn}
\affiliation{%
  \institution{Wangxuan Institute of Computer Technology, Peking University}
  \city{Beijing}
  \country{China}}

\author{Yang Liu}
\authornote{Corresponding author.}
\email{yangliu@pku.edu.cn}
\affiliation{%
  \institution{Wangxuan Institute of Computer Technology, Peking University}
  \city{Beijing}
  \country{China}}

\renewcommand{\shortauthors}{Gao et al.}

\begin{abstract}
Identity-preserving video generation aims to synthesize videos that follow natural-language instructions while maintaining the visual identity of a given subject. Recent commercial video generation models have achieved strong visual quality and motion realism, but they still suffer from identity drift, incomplete instruction following, and missing visual details under complex prompts. Since these models are usually closed-source black boxes, directly improving them through parameter optimization is often infeasible. We therefore propose Agentic Enhancement and Semantic Repair (AESR), a lightweight enhancement framework for identity-preserving video generation.
To improve prompt construction before generation and mitigate the above failures, AESR introduces a global agentic prompt enhancement module. This module learns model-specific prompting formats from official documentation, acquiring human-centered video generation priors from human-interaction data, and accumulates test-domain identity-preserving generation experience into a reusable playbook through an agentic loop. To further repair errors of videos generated with enhanced prompts, AESR further introduces a sample-level visual semantic repair module, which uses a VLM to locate erroneous video segments and design repair instructions, edits selected frames into explicit visual references, and guides a video editing model to fix local semantic or identity-related errors. We also adopt a lightweight Mixture-of-Experts selection strategy to choose reliable outputs from different generation and refinement paths. Under the official evaluation protocol of the ACM MM 2026 Identity-Preserving Video Generation Challenge, our system MIPL\_Video ranked first in Track~1, demonstrating the effectiveness of AESR for practical identity-preserving video generation. The code is available at \url{https://github.com/oceanflowlab/AESR}.
\end{abstract}

\ccsdesc[500]{Computing methodologies~Computer vision tasks}
\ccsdesc[300]{Computing methodologies~Image and video acquisition}

\keywords{text-to-video generation, identity preserved video generation, agentic based video generation}

\maketitle
\begin{figure*}[htbp]
    \centering
    \includegraphics[width=0.75\textwidth]{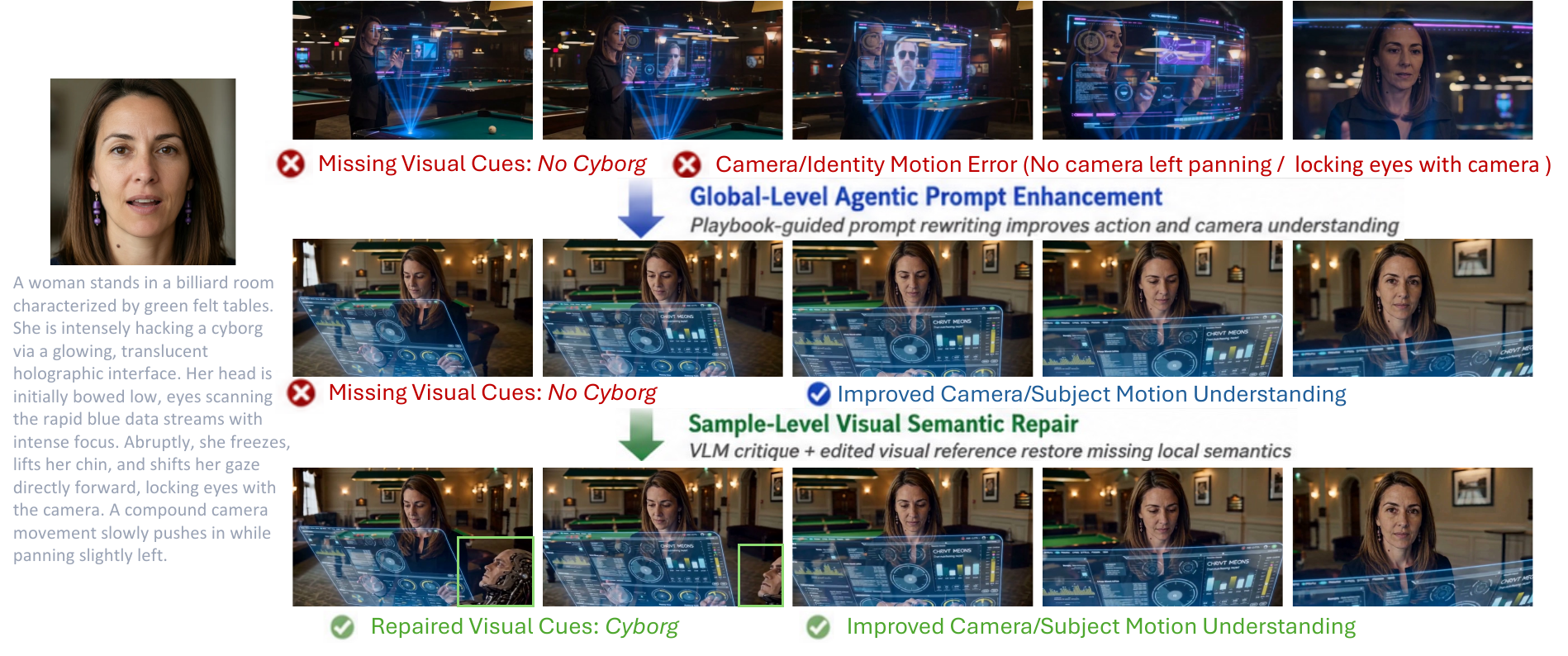}
    \caption{
AESR progressively improves identity-preserving video generation.
Global-level agentic prompt enhancement generally improves prompt alignment by improved motion understanding, while sample-level visual semantic repair restores missing visual elements.
}
    \label{fig:teaser}
\end{figure*}

\section{Introduction}
Identity-preserving video generation aims to synthesize a video that follows a natural-language instruction while maintaining the visual identity of a given source subject. Recent commercial video generation models, such as Seedance 2.0 \cite{seedance2026seedance}, have achieved impressive visual quality and motion realism. However, they still struggle in identity-sensitive scenarios: the generated subject may drift from the source identity, and complex instructions involving multiple visual elements or motion events may lead to missing objects, incomplete actions, or underspecified target states. Since these models are typically closed-source black boxes, directly improving them through parameter optimization is often infeasible. We therefore focus on what remains controllable at the interface level: prompt construction before generation and draft repair through external editing models afterward.

A straightforward way to mitigate these failures is to sample multiple candidates or manually rewrite the prompt. However, both solutions are costly and brittle. Repeated sampling requires multiple calls to expensive commercial APIs, while manual rewriting heavily depends on rewriter's experience. More importantly, effective prompts are shaped by both model-specific preferences and test-domain experience. Different video models may prefer different instruction formats due to their training data, such as how action decomposition, subject identity and camera motions are arranged. Meanwhile, test samples with similar instruction demands, such as complex actions or camera-motion control, often share reusable domain-level experience.
Moreover, some errors are difficult to fix by simply expanding the text prompt. Even when the instruction explicitly specifies certain visual elements, such as the cyborg in Fig. \ref{fig:teaser}, the generated video may still omit them. Since video models must coordinate appearance, actions, and camera motion across multiple frames, adding more text can introduce ambiguity when correcting local visual omissions; explicit visual references instead provide concrete guidance for repairing.

Therefore, we propose \textbf{AESR}, a lightweight enhancement framework for IPVG via \textbf{Agentic Enhancement and Semantic Repair}, consists of two parts: \textbf{Global‑level Prompt Enhancement} via an agentic loop, and \textbf{Sample‑level Visual Semantic Repair}. At the global level, inspired by \cite{zhang2025agentic}, we introduce an agentic loop mechanism that converts model prompt preferences and test‑domain experience into reusable generation knowledge and store them into an ever‑evolving playbook. 
The initial knowledge of the Playbook comes from two sources: the official prompting documentation of the target video model, and an external Playbook learned from the Human-object Interaction (HOI) Editing dataset \cite{gao2026taming}. The former provides model-specific prompting preferences, allowing the VLM to learn the instruction organization favored by the target model. The latter provides experience from HOI video generation. Since more than 95\% of the identity-preserving videos in the test set involve human interactions, as estimated using GPT-4o \cite{hurst2024gpt}, we draw prior knowledge from relevant external datasets to enhance interaction descriptions.
In the loop, the agent  analyzes failed test videos and incrementally updates the Playbook with generalizable experience, allowing it to adapt from initialized knowledge to the test domain. The updated Playbook is then used to enhance prompts for challenging videos that require stronger instruction following as in Fig \ref{fig:teaser}.
At the sample level, we introduce visual semantic repair to address errors that text alone cannot resolve. For visual errors in the initial video, such as identity drift or missing elements, edited image references provide concrete target states and serve as visual anchors for the motion-aware video editor to repair action and camera-motion errors. Specifically, we use a VLM \cite{team2023gemini,peng2026finegrained} to locate the approximate erroneous segment and design a repair instruction, and we use an image editor to correct specific frames within that segment as explicit visual references showing “what should appear”.These repaired reference frames, together with the original video, are then fed into a video editing model to supplement or fix the missing or incorrect visual elements in the generated result.

Putting these components together, we provide a practical enhancement solution for identity-preserving video generation with closed-source video models. In addition, we follow a lightweight Mixture-of-Experts (MoE) selection strategy to compare and integrate candidate videos produced by different generation or refinement paths, selecting the most reliable result for final submission. According to the official evaluation protocol of the ACM MM 2026 Identity-Preserving Video Generation Challenge, our system ranked first on the Facial Identity-Preserving Video Generation Track, validating the effectiveness of the proposed framework.

To sum up, our contributions are threefold:
\begin{itemize}
  \item We propose AESR, a lightweight interface-level enhancement framework for identity-preserving video generation with closed-source video models.

  \item We introduce two complementary modules in AESR: a global agentic prompt enhancement module that converts model-specific prompt preferences and test-domain experience into reusable generation knowledge for enhancing input prompts, and a sample-level visual semantic repair module that localizes errors, builds edited references, and uses video editing models for targeted refinement.

  \item Under the comprehensive official evaluation protocol of the ACM MM 2026 Identity-Preserving Video Generation Challenge, our system ranked first in the Facial Identity-Preserving Video Generation Track.
\end{itemize}


\section{Related Work}

Diffusion models \cite{esser2024scaling, yang2024cogvideox, kong2024hunyuanvideo,xie2025textvideo,cai2026music,xu2025dynamic,gao2025conmo} have propelled significant progress in many downstream tasks \cite{magictime, controlnet, InstaDrag, evagaussians, cycle3d, ViewCrafter} including identity-preserving generation \cite{dreamvideo, customvideo,xu2025psanerf}. Early approaches primarily relied on per-ID fine-tuning methods, such as MotionBooth \cite{motionbooth} and DreamVideo \cite{dreamvideo}, which incorporated reference content by fine-tuning model parameters or introducing additional modules. However, these methods required retraining for each new identity, greatly limiting scalability and practical deployment. To address these challenges, tuning-free strategies emerged, ACE++ \cite{mao2025ace++} and PhotoMaker \cite{photomaker} developed subject-preserved image generation models based on this approach. 
More recently, advanced models like Phantom \cite{liu2025phantom}, VACE \cite{jiang2025vace} and LTX-2 \cite{hacohen2026ltx} have demonstrated the capability to generate consistent multi-subject videos in open-domain scenarios \cite{liang2025movie, chen2025multi}, steadily closing the performance gap with commercial solutions like Hailuo \cite{hailuo2024}, Vidu \cite{Vidu} and Seedance 2.0\cite{seedance2026seedance}. Nevertheless, these methods require collecting large amounts of data and time-consuming post-training. 
Instead, we treat identity-conditioning mechanism as an implementation choice and study how to improve identity-preserving video generation through interface-level prompt enhancement and post-generation repair, which is more compatible with closed-source video models.
\begin{figure*}[htbp]
    \centering
    \includegraphics[width=0.8\textwidth]{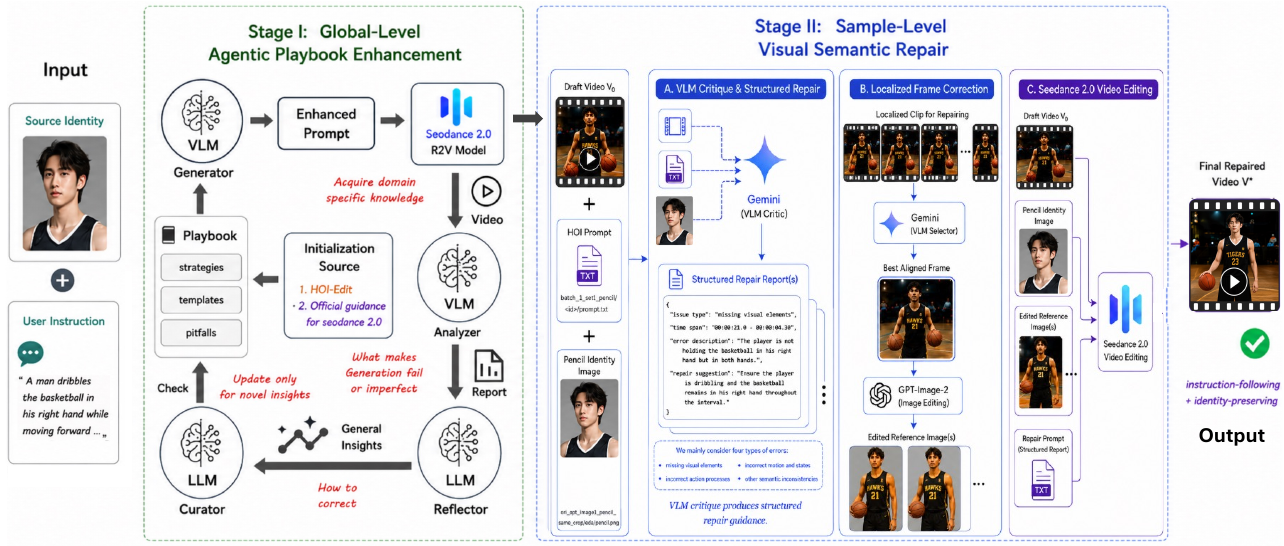}
    \caption{Two-stage AESR pipeline: Stage I shows global-level Agentic Playbook Enhancement, Stage II shows sample-level Visual Semantic Repair with pencil-style reference images.}
    \label{fig:aesr_pipeline}
\end{figure*}

\section{Method}

\subsection{Overview}
Given a source identity image I and a user instruction P, AESR aims to generate a video that faithfully follows the instruction while preserving the visual identity of the source subject. Since modern video generation and editing models are typically black-box systems, AESR improves controllability through two interface-level interventions. As illustrated in Fig. \ref{fig:aesr_pipeline}, to acquire and exploit generalizable knowledge for enhancing the instruction of samples, we propose \textbf{Global-Level Agentic Playbook Enhancement}, which aims to convert model-specific prompting preferences, human-interaction video generation experience, and failure experience observed from test samples into reusable generation knowledge, thereby improving the quality of generated videos. To address sample-specific errors that still remain in the videos produced by Stage I, we further propose \textbf{Sample-Level Visual Semantic Repair}, which focuses on diagnosing and correcting semantic gaps, missing visual elements, and identity-related errors in individual generated videos. 



\subsection{Global-Level Agentic Playbook Enhancement}

To continuously acquire and reuse generalizable generation knowledge from the test domain, we propose Global-Level Agentic Playbook Enhancement. The key motivation is that many failures in identity-preserving video generation are not isolated, but instead exhibit transferable patterns across different test samples. For example, complex actions often require clearer temporal decomposition and identity preservation requires more fine-grained appearance descriptions. Therefore, rather than manually rewriting the prompt for each individual sample, AESR aims to extract reusable experience from test-time generation feedback and convert it into general knowledge that can benefit subsequent samples.

Specifically, the Playbook stores reusable knowledge for prompt enhancement, including high-level strategies, general templates, and common failure modes. Strategies describe general generation principles, such as staged action descriptions, strengthened identity details, and the separation of subject motion from camera motion. Templates define how different prompt components should be organized, such as the ordering of human actions, human-object interactions, and target-state descriptions. Failure modes record common errors of the target model and their remedies, such as missing objects, weak identity constraints, or unclear final states. With this structured organization, the Playbook remains compact while providing clear guidance for subsequent prompt enhancement.


AESR initializes the Playbook from two complementary sources. The first source is the official prompting documentation of the target video model, which provides model-specific priors about how to organize identity, action, scene, style, and camera descriptions. By analyzing these documents, a VLM automatically learns the instruction organization favored by the target model and assigns the extracted experience into the three types of Playbook knowledge, helping the Playbook adapt to the input preferences of the target model. The second source is an external Playbook with the same format, learned from the HOI-Edit human-object interaction dataset and directly merged into the initial Playbook. Since identity-preserving video generation contains many human interaction events, such as contact, manipulation, movement, and state changes between humans and objects, we aim to draw reusable experience from relevant external human-object interaction data. 

To enable the Playbook to accumulate test-domain-specific guidance from cross-sample feedback, AESR updates the initialized Playbook through an agentic loop using the generation results of the test set. The loop consists of four steps: generation, analysis, reflection, and consolidation. First, the generator enhances the input instruction with the current Playbook and produces a draft video. Second, the analyzer inspects the draft video and identifies its failures. Third, the reflector abstracts these concrete failures into reusable experience, such as stronger object constraints, clearer temporal staging, or more detailed identity descriptions. Finally, the curator reviews the newly extracted experience together with existing Playbook knowledge and incrementally writes only useful, non-redundant guidance back into the Playbook. In this way, prompt enhancement can automatically accumulate and reuse knowledge from feedback across test samples.

Finally, the updated Playbook is used to enhance prompts for challenging samples. The agentic prompt generator incorporates relevant strategies, templates, and pitfalls to produce an enhanced prompt better suited to the target model, which is then used to generate the initial draft video.

\subsection{Sample-Level Visual Semantic Repair}

Although global prompt enhancement improves the initial draft video, some errors are highly sample-specific and cannot be reliably fixed by adding more textual descriptions. For example, the generated video may miss visual elements required by the instruction,  drift from the source identity, or misunderstand camera motions. To address these errors, AESR introduces Sample-Level Visual Semantic Repair, which uses visual diagnosis and explicit edited reference frames to guide video editing for repairing.

First, AESR uses a vision-language critic to compare the draft video with the original instruction and identify their mismatches. The critic produces a structured diagnosis, including the error type, problematic temporal segment, error description and repair suggestion. We mainly consider four types of errors: missing visual elements, incorrect motion end states, incorrect motion processes, and other semantic inconsistencies. Errors related to missing elements or incorrect end states are usually repaired with edited keyframes, while process-level errors, such as camera moving speed, or transition patterns, are repaired with textual editing instructions.

Second, for errors that require visual guidance, AESR selects a representative keyframe from the problematic segment and edits it according to the repair suggestion. The edited frame serves as an explicit visual target for the repaired video. This is useful when text alone is insufficient to describe spatial relations, object appearance, human pose, or identity-related details.

Finally, AESR constructs the repair inputs for Seedance 2.0 video editing, including a comprehensive editing instruction organized by the VLM \cite{team2023gemini}, the edited keyframes corresponding to the target temporal segments, and the original identity reference image. The comprehensive instruction is generated based on the structured diagnosis, the time-annotated editing requirements, and the original video prompt. Seedance 2.0 then edits the draft video according to these inputs and produces the final repaired result.

\section{Experiments}

\begin{figure*}[htbp]
    \centering
    \includegraphics[width=0.8\textwidth]{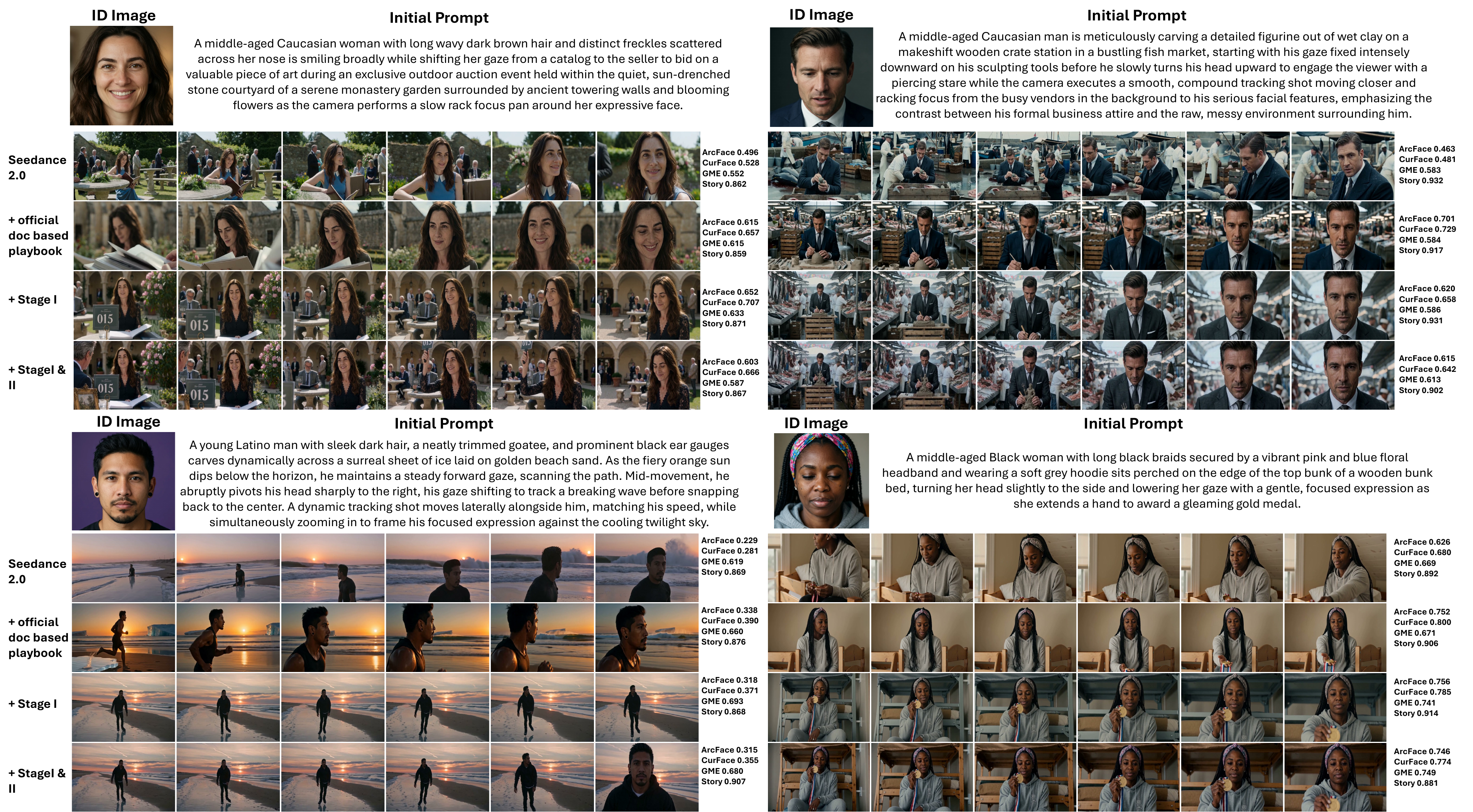}
    \caption{
Qualitative Results.
}
    \label{fig:vis}
\end{figure*}

\subsection{Challenge Setting and Evaluation Metrics}



Track 1 of the Identity-Preserving Video Generation (IPVG) Challenge, namely Facial IPVG, aims to generate videos from textual prompts while consistently preserving the facial identity of a given reference subject throughout the generated video. The Facial IPVG test set contains 200 evaluation samples, each consisting of one reference image and one textual prompt for video generation. The official evaluation assesses generated videos from two primary aspects: identity preservation and video quality. Identity preservation measures whether the generated video maintains the facial characteristics of the reference identity across the entire video, while video quality evaluates perceptual aspects including visual fidelity, motion smoothness, and text alignment through a combination of automatic metrics and human evaluation. Detailed information about the challenge can be found on the official website.\footnote{\url{https://hidream-ai.github.io/ipvg-challenge-2026.github.io/}}

To obtain a more comprehensive assessment of generated videos and support the subsequent MoE-based selection strategy, we further evaluate each video from three perspectives: text alignment, identity consistency, and video quality.
1. For text alignment, previous methods based on CLIP~\cite{radford2021learning} suffer from limitations such as the maximum token length of 77. In our evaluation, 89\% of the textual prompts are truncated when computing CLIPScore. Therefore, we do not adopt CLIPScore as a primary evaluation metric. Instead, we use StoryScore~\cite{shi2026msvbench} to measure story-level semantic consistency between the generated video and the prompt. Specifically, dense video captions are first generated and then compared with the original prompt based on semantic similarity. We further adopt GMEScore~\cite{zhang2024gme} as a complementary text-alignment metric.
2. For identity consistency, we calculate the similarity between generated video frames and the reference image using two facial recognition models, namely ArcFace~\cite{arcface} and CurricularFace~\cite{curricularface}, resulting in ArcScore and CurScore, respectively.
3. For video quality, we adopt two metrics from VBench~\cite{huang2024vbench}, including Motion Smoothness and Imaging Quality, to evaluate motion quality and visual quality.
All adopted metrics are normalized to the range of $[0,1]$, where higher values indicate better performance.

\subsection{MoE Selection Strategy}
\label{sec:moe}

Different methods exhibit complementary strengths in text alignment, identity preservation, and video quality. To fully leverage these advantages and achieve the best challenge performance, we adopt a Mixture-of-Experts (MoE) strategy to select the optimal result for each sample.
For each generated video, we first compute an overall score based on the evaluation metrics described above:
\begin{equation}
\text{Overall Score}
=
\sum_{i \in \mathcal{M}}
w_i \cdot M_i,
\end{equation}
where $\mathcal{M}$ denotes the set of evaluation metrics, $M_i$ is the score of the corresponding metric, and $w_i$ is its associated weight.
For each test sample, we generate candidate videos using five different methods and calculate the overall score for each candidate. The video with the highest overall score is the final submission result.


\subsection{Implementation Details}

We use Seedance 2.0 as the main closed-source video generation and editing backbone. For each sample, the input consists of a source identity image and a natural-language instruction. Before calling the video APIs, we convert all identity-bearing images into pencil-style identity proxies. This representation preserves identity-relevant facial structure and shape while reducing the influence of irrelevant appearance factors such as background color and illumination, supporting stable batch API calls. This format is used consistently in both Stage I reference-to-video generation and Stage II visual semantic repair with edited image references. We generate 14-second videos for all samples, since the prompts often contain multiple visual elements and motion events; the longer duration gives the model more temporal capacity to cover the requested content.
For Stage I, we first generate an initial round of videos using the original prompts from the test set, and analyze these results with the initialized Playbook to extract reusable experience. We then further include failure cases produced by different variants of our method in the Playbook update process, allowing it to accumulate test-domain knowledge from richer cross-sample feedback.

\subsection{Qualitative Results}
Figure~\ref{fig:vis} shows qualitative results of different AESR variants, including vanilla Seedance generation, prompt enhancement with playbook initialized with only official documentation, Stage-I global enhancement, and the final result with Stage-II visual semantic repair. The results show that introducing playbook knowledge improves alignment with both textual content and motion descriptions in the prompt. For example, in the bottom-left case, the prompt requires the subject to maintain a steady forward gaze while scanning the path, which is better captured after playbook enhancement. It also reduces generation artifacts, like the duplicated identity subject in the top-right case.
The comparison between the second and third rows further shows the importance of incorporating test-domain experience into GPE. Compared with using only official prompting formats, domain experience better preserves complex actions and visual details. For example, the bottom-left case better reflects the action and scene relation described by ``carves dynamically across a surreal sheet of ice''; the top-right case more accurately realizes the behavior of staring at the viewer; and the bottom-right case more reliably generates the required bunk-bed scene. This indicates that GPE not only learns model-preferred prompt formats, but also accumulates domain-level knowledge useful for complex instruction following.
In Stage II, SVR further corrects local visual semantic errors that remain in the draft video. Although the draft may preserve the overall scene and motion structure, it can still miss key visual cues, subject/camera motion, or identity details. For example, the top-left case may miss the gaze transition required by ``shifting her gaze from a catalog to the seller to bid''; the bottom-left case may not clearly reach the final camera state of ``zooming in to frame his focused expression''; and the bottom-right case may drift on identity details such as ``a vibrant pink and blue floral headband''. SVR uses a VLM to locate problematic segments and design repair instructions, edits selected frames as explicit visual references, and guides the video editing model to repair incorrect local elements.
These qualitative results show that global prompt enhancement and sample-wise visual repair are complementary: the former improves draft generation at the instruction level, while the latter provides direct visual guidance for fine-grained local errors.

\begin{table*}[t]
\centering
\caption{Average metrics on the selected 50 videos from Testset of IPVG2026.}
\label{tab:component_analysis}
\resizebox{\textwidth}{!}{%
\begin{tabular}{lcc|cc|ccc|cc}
\toprule
\multirow{2}{*}{Method} 
& \multicolumn{2}{c|}{Text Alignment} 
& \multicolumn{2}{c|}{Identity Consistency} 
& \multicolumn{3}{c|}{Video Quality} 
& \multirow{2}{*}{OverallScore$\uparrow$}
& \multirow{2}{*}{HumanScore$\uparrow$} \\
\cmidrule(lr){2-3} \cmidrule(lr){4-5} \cmidrule(lr){6-8}
& StoryScore$\uparrow$ 
& GMEScore$\uparrow$ 
& CurScore$\uparrow$ 
& ArcScore$\uparrow$ 
& Motion$\uparrow$ 
& Imaging$\uparrow$ 
& FID$\downarrow$ 
& & \\
\midrule
Vanilla generation 
& 0.8965 & 0.6563 & 0.5165 & 0.4755 & 0.9933 & 0.6662 & 220.11 & 0.6803 & 3.42 \\

Official prompt enhancement 
& 0.8916 & 0.6608 & \textbf{0.6253} & \textbf{0.5868} & 0.9938 & 0.6814 & \textbf{166.35} & \textbf{0.7266} & 3.68 \\

+ GPE 
& 0.8972 & 0.6688 & 0.5952 & 0.5618 & \textbf{0.9939} & 0.6836 & 187.76 & 0.7179 & 3.82 \\

+ text-only repair 
& 0.8952 & 0.6704 & 0.6082 & 0.5732 & 0.9929 & \textbf{0.6961} & 175.25 & 0.7245 & 3.74 \\

Full AESR 
& \textbf{0.8994} & \textbf{0.6776} & 0.5762 & 0.5441 & 0.9934 & 0.6885 & 184.89 & 0.7129 & \textbf{4.01} \\
\bottomrule
\end{tabular}%
}
\end{table*}

\begin{table*}[t]
\centering
\caption{Comparison with existing identity-preserving video generation methods.}
\label{tab:main_comparison}
\resizebox{\textwidth}{!}{%
\begin{tabular}{lcc|cc|ccc|c}
\toprule
\multirow{2}{*}{Methods} 
& \multicolumn{2}{c|}{Text Alignment} 
& \multicolumn{2}{c|}{Identity Consistency} 
& \multicolumn{3}{c|}{Video Quality} 
& \multirow{2}{*}{OverallScore$\uparrow$} \\
\cline{2-8}
& CLIPScore$\uparrow$ 
& GMEScore$\uparrow$ 
& CurScore$\uparrow$ 
& ArcScore$\uparrow$ 
& Motion$\uparrow$ 
& Imaging$\uparrow$ 
& FID$\downarrow$ 
& \\
\midrule
Hailuo~\cite{hailuo2024}
& 30.15 & 0.6087 & 0.0866 & 0.0705 
& \textbf{0.9880} & \textbf{0.6772} & 243.54 & 0.4638 \\

Phantom-14B~\cite{liu2025phantom}
& 30.17 & 0.6231 & 0.2459 & 0.2391 
& 0.9810 & 0.6367 & 268.61 & 0.5266 \\

VACE-14B~\cite{jiang2025vace}
& 30.00 & 0.6187 & 0.2034 & 0.1927 
& 0.9750 & 0.6349 & 250.35 & 0.5063 \\

TPIGE~\cite{gao2025identity}
& 29.04 & 0.6040 & 0.2564 & 0.2424
& 0.9701 & 0.6265 & 252.88 & 0.5204 \\

Seedance2 \cite{seedance2026seedance}
& 30.59 & 0.6560 & 0.2189 & 0.1997 
& 0.9789 & 0.6348 & 265.95 & 0.5226 \\

Seedance2 + HOI-Edit based playbook \cite{seedance2026seedance,gao2026taming}
& \textbf{31.26} & \textbf{0.6598} & \textbf{0.2834} & \textbf{0.2662} 
& 0.9824 & 0.6548 & \textbf{239.81} & \textbf{0.5534} \\
\bottomrule
\end{tabular}%
}
\end{table*}


\begin{table}[t]
\centering
\caption{Official leaderboard of Track 1 -- Facial Identity-Preserving Video Generation in the ACM MM 2026 Identity-Preserving Video Generation Challenge.}
\label{tab:official_leaderboard}
\resizebox{0.5\columnwidth}{!}{%
\begin{tabular}{clc}
\toprule
Rank & Team & Final Score$\downarrow$ \\
\midrule
1 & MIPL\_Video & 2.5 \\
1 & USTC-CMI & 2.5 \\
3 & xuanyuan\_fuxi & 3.0 \\
\bottomrule
\end{tabular}%
}
\end{table}

\subsection{Quantitative Results}

Before the release of the official IPVG 2026 test set, we first compared different base video generation models on an earlier IPVG 2025 evaluation set, and also tested the enhancement strategy based on the HOI-Edit Playbook. This evaluation was used for model selection before this year's official test results became available. As shown in Table~\ref{tab:main_comparison}, Seedance2 achieves the strongest text alignment and a competitive overall score among existing identity-preserving video generation methods. Since the new test set contains more complex textual descriptions, we adopt Seedance2 as the base generation and editing model for AESR. Furthermore, the HOI-Edit-based Playbook initialization strategy further improves the overall performance, demonstrating that human-object interaction experience can effectively enhance the quality of identity-preserving video generation on this benchmark.
We then analyze AESR components on 50 selected videos from the IPVG 2026 test set in Table~\ref{tab:component_analysis}. Different paths show complementary strengths. Official prompt enhancement achieves the best CurScore and ArcScore, suggesting that model-provided prompting priors help identity-related metrics. GPE further improves StoryScore and GMEScore, showing that test-domain Playbook knowledge helps complex instruction following. Text-only repair improves Imaging Quality, while full AESR obtains the best text-alignment scores.

The component results also reveal limitations of current automatic metrics. Face-recognition metrics such as ArcScore can fluctuate noticeably even when identities look well preserved to humans. As shown in the top-right example of Fig.~\ref{fig:vis}, the second and third rows look similar in identity preservation but have clearly different face scores. In contrast, text-alignment metrics more directly reflect whether complex instructions are followed, but their score range is smaller and thus less visible in the overall score.
Therefore, following the Track~1 evaluation protocol of the ACM MM 2026 Identity-Preserving Video Generation Challenge, we further conduct human scoring that jointly considers identity preservation and video quality. Based on this human evaluation, full AESR achieves the best performance among the compared variants. The final leaderboard further supports our findings. As shown in Table~\ref{tab:official_leaderboard}, under the official protocol combining automatic metrics and human evaluation, our team MIPL\_Video with AESR ranked first in Track~1 Facial Identity-Preserving Video Generation with a final score of 2.5, demonstrating the effectiveness of the full system.

\section{Conclusion}


We presented AESR, a lightweight interface-level framework for identity-preserving video generation with closed-source video models. AESR improves controllability without accessing model parameters by combining global agentic prompt enhancement with sample-level visual semantic repair. The former accumulates reusable generation knowledge from prompting priors and test-sample feedback, while the latter uses visual diagnosis and edited reference frames to correct local semantic and identity-related errors. A lightweight MoE strategy further selects reliable outputs across different generation and refinement paths. Extensive experiments demonstrate that AESR consistently improves identity preservation, semantic fidelity, and visual quality across diverse editing scenarios. Furthermore, our solution achieved first place in Track 1 of the official ACM MM 2026 Identity-Preserving Video Generation Challenge, validating the effectiveness and practical applicability of AESR.

\noindent\textbf{Acknowledgements.}This work was supported by the grants from the National Natural Science Foundation of China (62372014, 62525201,
62132001, 62432001), Beijing Nova Program and Beijing
Natural Science Foundation (4252040, L247006).

\bibliographystyle{ACM-Reference-Format}
\bibliography{references}

\end{document}